\documentclass[10pt,twocolumn,letterpaper]{article}

\usepackage{cvpr}              

\usepackage{subcaption}

\definecolor{cvprblue}{rgb}{0.21,0.49,0.74}
\usepackage[pagebackref,breaklinks,colorlinks,allcolors=cvprblue]{hyperref}

\def\paperID{22797} 
\def\confName{CVPR}
\def\confYear{2026}

\title{CHNNet: An Artificial Neural Network With Connected Hidden Neurons}

\author{Rafiad Sadat Shahir
\and
Zayed Humayun
\and
Mashrufa Akter Tamim
\and
Shouri Saha
\and
Md. Golam Rabiul Alam
\and
Abu Mohammad Khan
\and
BRAC University\\
{\tt\small {\{rafiad.shahir,zayed.humayun,rabiul.alam,abu.khan\}@bracu.ac.bd}}\\
{\tt\small {\{mashrufa.akter.tamim,shouri.saha\}@g.bracu.ac.bd}}
}

\begin{document}
\maketitle
\begin{abstract}
In contrast to biological neural circuits, conventional artificial neural networks are commonly organized as strictly hierarchical architectures that exclude direct connections among neurons within the same layer. Consequently, information flow is primarily confined to feedforward and feedback pathways across layers, which limits lateral interactions and constrains the potential for intra-layer information integration. We introduce an artificial neural network featuring intra-layer connections among hidden neurons to overcome this limitation. Owing to the proposed method for facilitating intra-layer connections, the model is theoretically anticipated to achieve faster convergence compared to conventional feedforward neural networks. The experimental findings provide further validation of the theoretical analysis.
\end{abstract}
\section{Introduction}
Artificial Neural Networks (ANNs) are conceptualized as simplified computational models inspired by the organizational principles of biological neural systems. Empirical studies suggest that the human brain exhibits dense hierarchical and lateral connectivity patterns, which facilitates dynamic information exchange and adaptive processing~\cite{cri93,gro87}. Such lateral interactions among neural populations play a crucial role in coordinating representations, minimizing redundancy, and enhancing the stability of learning processes. Early computational abstractions, such as the McCulloch–Pitts neuron~\cite{mcc43}, were inspired by biological principles but did not adequately represent the intricate connectivity patterns observed in real neural circuits.

As the field progressed, artificial neural architectures predominantly adopted a structured feedforward topology. In a typical Feedforward Neural Network (FNN), neurons in a given layer connect exclusively to neurons in adjacent layers, remaining independent of one another within the same layer. This design promotes computational efficiency and stable gradient propagation, but diverges from the biological principle of lateral interactivity. Consequently, while such architectures are effective in hierarchical information processing, they lack internal communication mechanisms that can potentially enable richer and more adaptive representational learning.

Several recent studies have extended beyond conventional designs to incorporate increased connectivity among the neurons~\cite{liu22,che20,zha21,che23,zha22,tho17,resnet}. Nonetheless, the introduction of intra-layer connections among hidden neurons and their potential impact on the network’s learning dynamics remain largely unexplored.

In exploring this issue, we introduce an extended feedforward model in which neurons within the same hidden layer share information through learnable lateral connections. This structural modification maintains the feedforward architecture while introducing intra-layer communication reminiscent of cortical interactions in biological systems. We refer to this network as CHNNet, with the primary aim of investigating the functional consequences of introducing lateral connectivity absent in conventional architectures.

Collectively, we reconsider a fundamental architectural assumption of feedforward networks that neurons within the same layer operate in isolation and demonstrate that relaxing this constraint can lead to notable improvements in learning dynamics.

\label{theoretical_study}
\begin{figure*}[t]
    \centering
    \begin{minipage}{0.75\columnwidth}
        \centering
        \includegraphics[width=\linewidth]{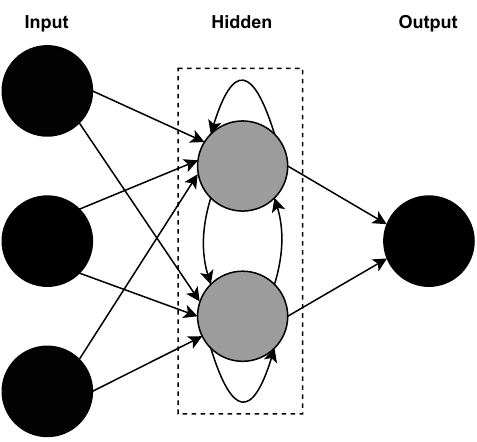}
        \small (a)
        \label{fig:img1}
    \end{minipage}
    \hspace{5pt}
    \begin{minipage}{0.9\columnwidth}
        \centering
        \includegraphics[width=\linewidth]{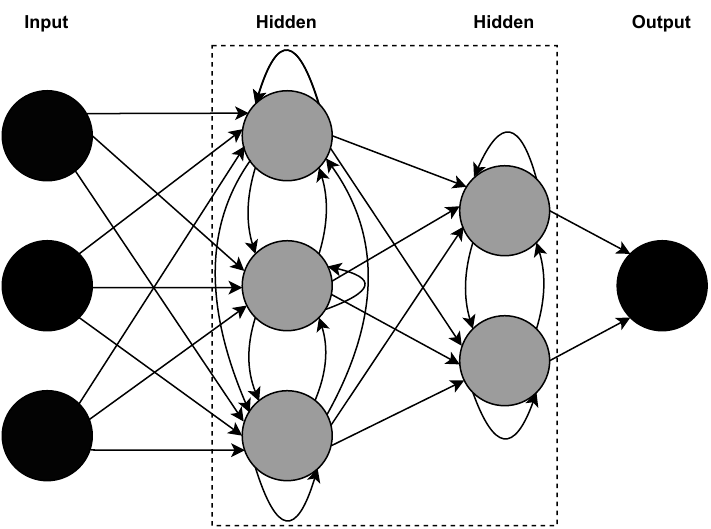}
        \small (b)
        \label{fig:img2}
    \end{minipage}
    \caption{The architecture of CHNNet with (a) one hidden layer and (b) two hidden layers.}
    \label{architecture}
\end{figure*}

\section{Theoretical Study}
In this study, we analyze the proposed CHNNet architecture from the perspective of conventional FNNs, focusing on its modifications to the hidden layer connectivity and their impact on the learning behavior of the network.

\subsection{Architecture}
The architecture of CHNNet extends conventional FNN architectures by incorporating additional interconnections among hidden neurons, as illustrated in figure~\ref{architecture} with example networks containing one and two hidden layers. We propose modifications exclusively to the connectivity of the hidden layers, while the input and output layers remain unaltered.

\subsection{Inference Mechanism}
We define the pre-activation output of a hidden layer in CHNNet, denoted by $z$, as follows:
\begin{equation}
\label{inf1}
    z^{[l]} = W_1^{[l]} a^{[l-1]} + W_2^{[l]} h^{[l]} + b^{[l]}
\end{equation}
Consider that $a$ is the post-activation output of a hidden layer, $h$ is the post-activation output of the hidden neurons of the current layer, $W_1$ is the weight matrix for the connections of the previous layer to the current layer, $W_2$ is the weight matrix for the intra-layer connections of the hidden layer, $b$ is the bias vector and $l$ denotes the index of the hidden layer.

The variable $h$ in equation (\ref{inf1}) is a distinctive component of CHNNet, differentiating it from conventional FNNs. We propose to compute $h$ by applying the indentity function $f_I$ as follows:
\begin{equation}
\begin{aligned}[b]
\label{inf2}
h^{[l]} &= f_I(W_1^{[l]} a^{[l-1]} + b^{[l]})\\
        &=W_1^{[l]} a^{[l-1]} + b^{[l]}
\end{aligned}
\end{equation}

Subsequently, the post-activation output of the hidden layer is computed by applying the activation function $f$ as follows:
\begin{equation}
    \label{inf3}
    a^{[l]} =f^{[l]}(z^{[l]})
\end{equation}

Since the connectivity of the output layer is unaltered, the post-activation output of the output layer for a network comprising $L$ layers is computed as follows:
\begin{equation}
    \label{inf4}
    z^{[L]} = W_1^{[L]} a^{[L-1]} + b^{[L]}
\end{equation}
\begin{equation}
    \label{inf5}
    a^{[L]} =f^{[L]}(z^{[L]})
\end{equation}

\subsection{Comparison with Multilayer Perceptrons}
In the foundational stages of neural network research, \citet{min69} introduced the conceptual basis of the Multilayer Perceptron (MLP), describing it as a hierarchical model without intra-layer connections among hidden neurons. Subsequently, \citet{rum86} conducted extensive analysis of various MLP architectures, none of which incorporated intra-layer connections. The pre-activation output in these studies is commonly formulated as follows:
\begin{equation}
    \label{lit1}
    z^{[l]} = W^{[l]} a^{[l-1]} + b^{[l]}
\end{equation}

Equation (\ref{lit1}) differs substantially from the corresponding CHNNet formulation in equation (\ref{inf1}).

\subsection{Comparison with Recurrent Neural Networks}
Recurrent Neural Networks~\cite{rum86} introduced feedback loops that allow information to propagate through time, enabling temporal modeling of sequential data. The Hopfield Network~\cite{hop82} and the Boltzmann Machine~\cite{hin86} further extended connectivity by employing symmetric bidirectional links among neurons, functioning respectively as an associative memory model and a probabilistic representation learner.

Neural architectures such as the Echo State Network~\cite{jae01} and the Liquid State Machine~\cite{wol02} employ a reservoir of randomly connected neurons that provides nonlinear modeling capabilities.

Within Spiking Neural Networks (SNNs), intra-layer connections in the hidden layer have been proposed by \citet{che20}, \citet{zha21}, and \citet{che23}, while self-connections have been introduced by \citet{zha22}.

Notably, each of the aforementioned studies incorporates some form of recurrent connectivity within the hidden neurons. In these studies, the pre-activation output at time step $t$ is generally formulated as follows:
\begin{equation}
\label{lit2}
z_t^{[l]} = W_1^{[l]} a_t^{[l-1]} + W_2^{[l]} h_{t-1}^{[l]} + b^{[l]}
\end{equation}

Nevertheless, the intra-layer connections in CHNNet differ fundamentally from recurrent connections. While RNNs compute the hidden layer output using activations from the previous time step, CHNNet utilizes the linear activations of hidden neurons within the current time step, as shown in equation (\ref{inf1}).

\subsection{Comparison with Skip-Connections}
Recent advances in neural network design have focused on creating new pathways for information flow. Convolutional Neural Network (CNN) architectures, including DenseNet~\cite{densenet}, ResNet~\cite{resnet}, UNet~\cite{unet}, and UNet++~\cite{unet++}, employ skip connections to propagate information directly from one layer to subsequent deeper layers.

Although the model’s computations of inputs from hidden neurons are sometimes interpreted as forming two conventional FNN layers linked by a skip connection, the architecture in fact implements intra-layer coupling within a single layer rather than a stacked two-layer structure. To clarify this distinction, consider the following derivation based on equation (\ref{inf1}):
\begin{equation}
\begin{aligned}[b]
\label{lit3}
    z &= W_1 a + W_2 h + b\\
            &= W_1 a + W_2 (W_1 a + b) + b\\
            &= W' a  + b'\\
\end{aligned}
\end{equation}

From equation (\ref{lit3}), it follows that the hidden layers of CHNNet operate effectively as a single layer, rather than as two distinct conventional FNN layers.

\subsection{Learning Mechanism}
In this study, we have analyzed CHNNet exclusively using backpropagation as the learning mechanism. CHNNet comprises three learnable parameters, specifically weights $W_1$, $W_2$, and bias $b$. Computing the gradients of these parameters is straightforward. 

For the $l^{th}$ layer in a conventional FNN, the partial derivative of the loss function, denoted as $J$, with respect to a learnable parameter $W$ is computed as follows:
\begin{equation}
    \label{lr1}
    \frac{\partial J}{\partial W^{[l]}} = 
    \frac{\partial J}{\partial z^{[l]}} \frac{\partial z^{[l]}}{\partial W^{[l]}}
\end{equation}

For CHNNet, the partial derivative of the loss function with respect to $W_2$ is computed as follows:
\begin{equation}
    \label{lr1}
    \frac{\partial J}{\partial W^{[l]}_2} = 
    \frac{\partial J}{\partial z^{[l]}} \frac{\partial z^{[l]}}{\partial W^{[l]}_2}
\end{equation}

Differing from conventional FNNs, the partial derivatives of the loss function with respect to $W_1$ and $b$ are computed as follows:
\begin{equation}
    \label{lr2}
    \frac{\partial J}{\partial W^{[l]}_1} = 
    \frac{\partial J}{\partial z^{[l]}} (\frac{\partial z^{[l]}}{\partial W^{[l]}_1} + \frac{\partial z^{[l]}}{\partial h^{[l]}} \frac{\partial h^{[l]}}{\partial W^{[l]}_1})
\end{equation}
\begin{equation}
    \label{lr3}
    \frac{\partial J}{\partial b^{[l]}} = 
    \frac{\partial J}{\partial z^{[l]}} (\frac{\partial z^{[l]}}{\partial b^{[l]}} + \frac{\partial z^{[l]}}{\partial h^{[l]}} \frac{\partial h^{[l]}}{\partial b^{[l]}})
\end{equation}

The changes in computation arise from the inclusion of the term $h$ in the post-activation output computation, where $h$ is computed using $W_1$ and $b$.

The procedure for computing the output layer gradients remains consistent with that of a conventional FNN.

\subsection{Convergence Analysis}
Upon analyzing the inference and learning mechanisms of CHNNet, we found that it is expected to converge more rapidly than a conventional FNN.

To simplify the analysis, we assume that the bias $b$ is a zero vector. Under this assumption, a conventional FNN employs a single learnable parameter $W_1$, while CHNNet employs two learnable parameters $W_1$ and $W_2$. Consider the local loss function $J_{c}$ associated with a hidden layer in CHNNet and $J_{f}$ associated with the corresponding hidden layer in a conventional FNN. At a given time step $t$ of the learning process, $J_{c}$ and $J_{f}$ are formulated as follows:
\begin{equation}
    \label{con1}
    J_{c}(W^{t}_1,W^{t}_2) = || o^{*} - f(z^t_c) ||
\end{equation}
\begin{equation}
    \label{con2}
    J_{f}(W^{t}_1) = || o^{*} - f(z^t_f) ||
\end{equation}

In equation (\ref{con1}), $z_c$ denotes the pre-activation output of a hidden layer of CHNNet, while in equation (\ref{con2}), $z_f$ denotes the pre-activation output of a hidden layer of a conventional FNN. In both equations, $o^{*}$ represents the optimal output of the hidden layer that the learning mechanism aims to determine.

Following equation (\ref{inf1}), the relation between $z_c$ and $z_f$ can be expressed as follows:
\begin{equation}
\begin{aligned}[b]
\label{con3}
    z_c &= z_f + W_2 h\\
        &= z_f + W_2 z_f\\
        &= (I + W_2) z_f
\end{aligned}
\end{equation}

To simplify the subsequent analysis, $W_2$ is assumed to be a zero matrix at time step $t-1$, which implies that $z_c^{t-1}$ is equal to $z_f^{t-1}$. Thus, at time step $t-1$ the following holds:
\begin{equation}
\begin{aligned}[b]
\label{con4}
    || o^{*} - f(z^{t-1}_c&) || = || o^{*} - f(z^{t-1}_f) ||\\
    \Rightarrow J_{c}(W^{t-1}_1,&W^{t-1}_2) = J_{f}(W^{t-1}_1)
\end{aligned}
\end{equation}

In accordance with the convergence theorem of gradient descent, $W_2$ is updated to ensure that $f(z_c)$ approaches $o^*$. Specifically, $W_2$ adjusts $z_f$ such that the corresponding output $f(z_c)$ progressively approaches $o^*$, yielding predictions that are consistently closer to $o^*$ than those produced by $f(z_f)$. Hence, at time step $t$, the following holds:
\begin{equation}
\begin{aligned}[b]
\label{con5}
    || o^{*} - f(z^t_c&) || < || o^{*} - f(z^t_f) ||\\
    \Rightarrow J_{c}(W^{t}_1,&W^{t}_2) < J_{f}(W^{t}_1)
\end{aligned}
\end{equation}

While the model has been analyzed using gradient-based methods, particularly backpropagation, equation (\ref{con5}) holds for any learning algorithm with theoretically guaranteed convergence.

Applying equation (\ref{con4}) in conjunction with inequality (\ref{con5}), we obtain the following:
\begin{equation}
\label{con6}
J_{c}(W^{t-1}_1,W^{t-1}_2) - J_{c}(W^{t}_1,W^{t}_2) > J_{f}(W^{t-1}_1) - J_{f}(W^{t}_1)
\end{equation}

Equation (\ref{con6}) indicates that the change in the loss of CHNNet between two successive time steps is greater than that of a conventional FNN. Consequently, CHNNet produces a steeper gradient, thus achieving faster convergence compared to conventional FNN.

\section{Empirical Study}
We empirically evaluated CHNNet on widely used architectures, including multilayer perceptrons (MLPs) and deep convolutional neural networks (CNNs). Since CHNNet inherently introduces additional parameters to implement intra-layer connections, we assessed its performance under equal-parameter conditions as well.

The models were trained using identical hyperparameters derived from the work of \citet{adam}. To ensure experimental fairness, all models were trained using identical weight initializations and fixed random seeds.

\begin{figure*}[t]
    \centering
    \begin{subfigure}{0.45\textwidth}
        \centering
        \includegraphics[width=\linewidth]{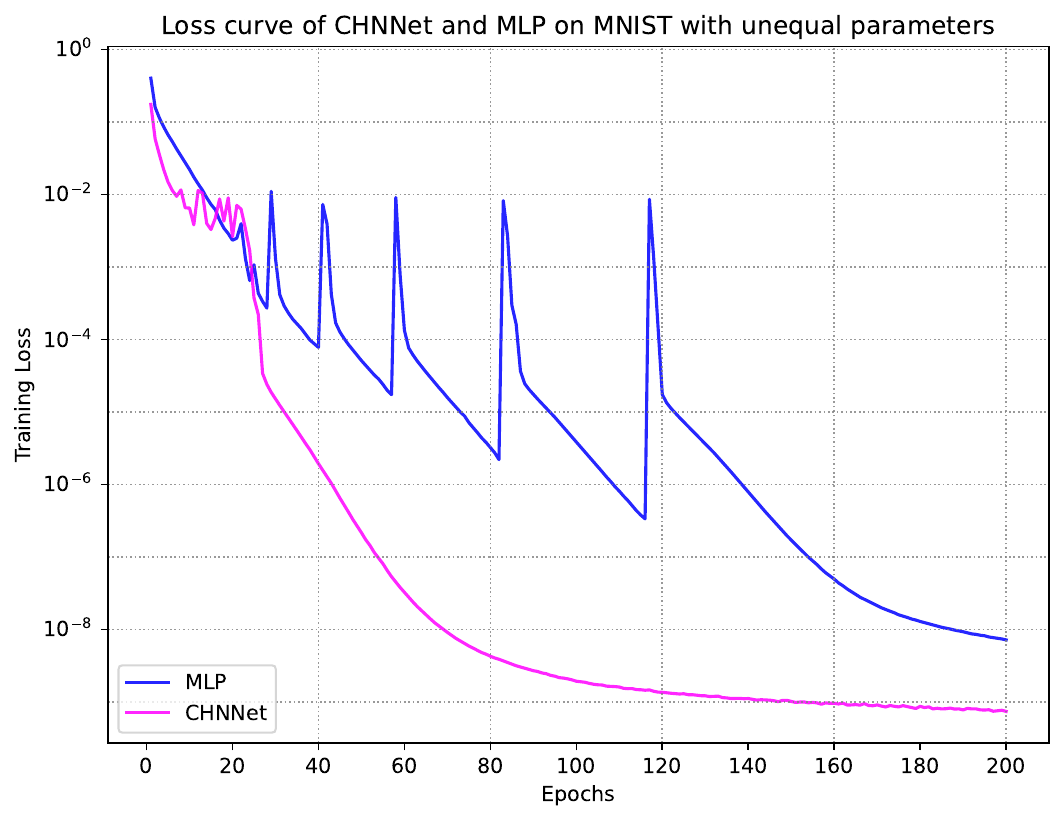}
        \caption{}
        \label{fig:img3}
    \end{subfigure}
    \hspace{5pt}
    \begin{subfigure}{0.45\textwidth}
        \centering
        \includegraphics[width=\linewidth]{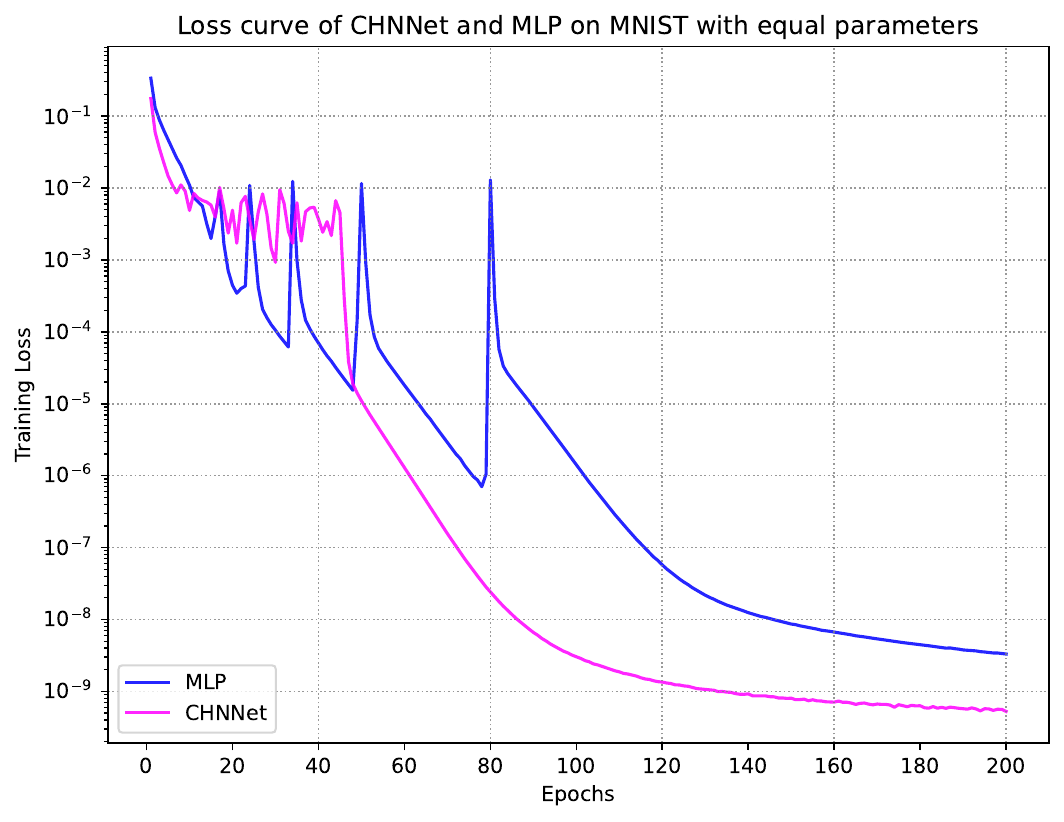}
        \caption{}
        \label{fig:img4}
    \end{subfigure}

    \vspace{6pt}

    \begin{subfigure}{0.45\textwidth}
        \centering
        \includegraphics[width=\linewidth]{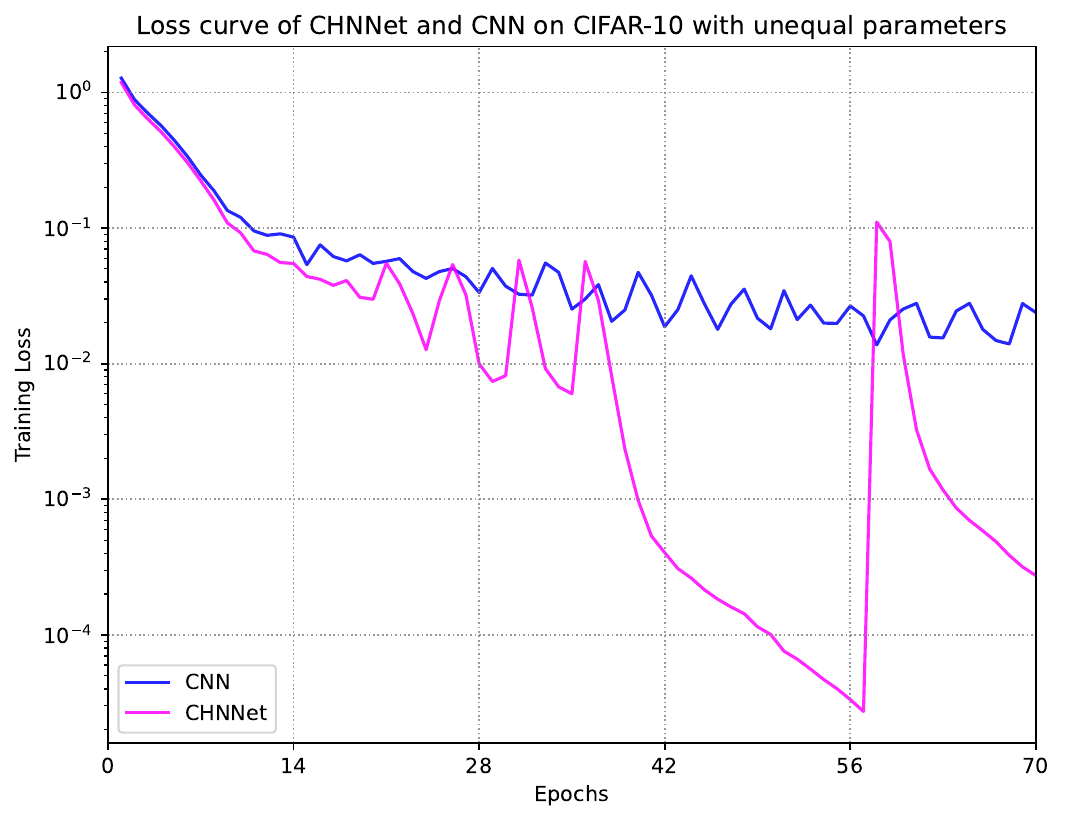}
        \caption{}
        \label{fig:img5}
    \end{subfigure}
    \hspace{5pt}
    \begin{subfigure}{0.45\textwidth}
        \centering
        \includegraphics[width=\linewidth]{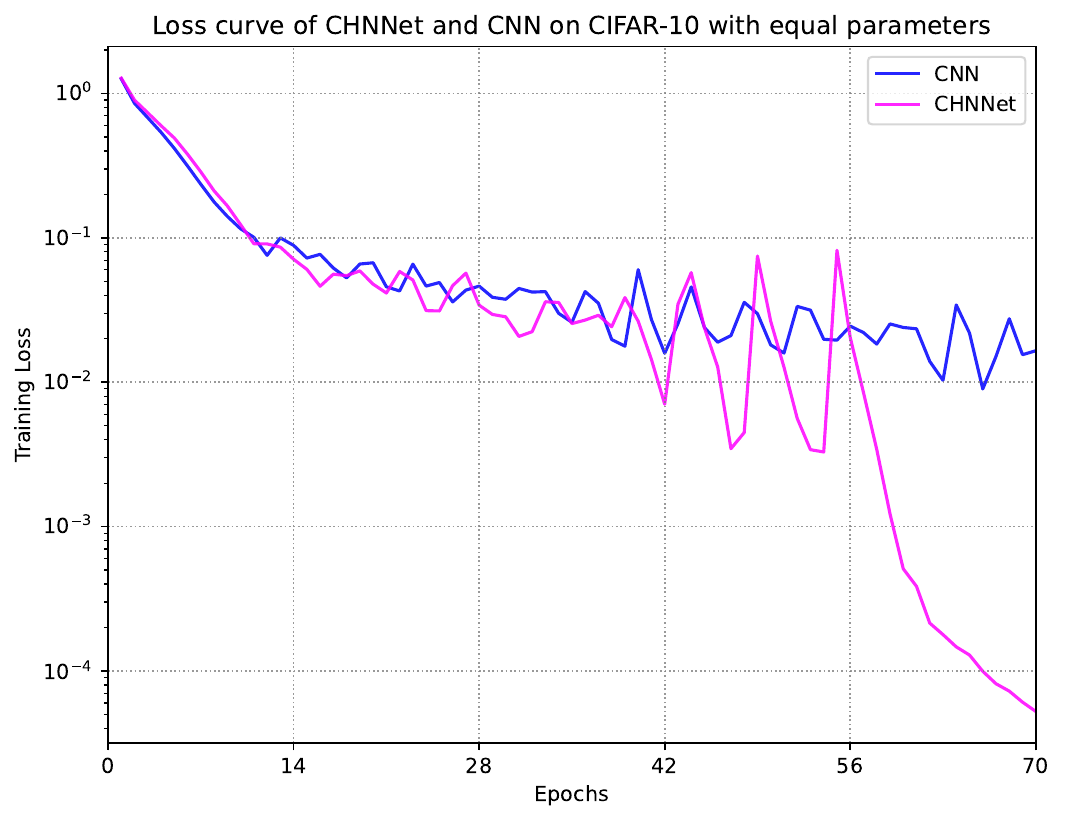}
        \caption{}
        \label{fig:img6}
    \end{subfigure}

    \caption{Loss curves on (a–b) MNIST and (c–d) CIFAR‑10 datasets. (a) MLP and CHNNet with unequal parameters, (b) MLP and CHNNet with equal parameters, (c) CNN and CHNNet with unequal parameters, (d) CNN and CHNNet with equal parameters.}
    \label{fig:mlp_cnn_chnnet}
\end{figure*}

\subsection{Experiment with MLP Architectures}
We evaluated the convergence behavior of the CHNNet architecture in comparison to a conventional MLP architecture on the MNIST dataset. In the unequal parameter setting, the MLP comprised two fully connected layers with 1000 units each, resulting in 1.7 million parameters. In contrast, CHNNet, with two hidden layers of the same size but incorporating intra-layer connections, comprised 3.7 million parameters. As visualized in Figure~\ref{fig:mlp_cnn_chnnet}\subref{fig:img3}, CHNNet exhibited a faster convergence than MLP within 200 epochs.

To ensure a balanced comparison, additional experiments were conducted with approximately equal parameter counts. In this setting, the MLP was widened to fully connected layers with 1,580 units each, resulting in roughly 3.7 million parameters, closely matching the 3.7 million parameters of CHNNet. Despite achieving near parity in parameter count, CHNNet consistently demonstrated faster convergence, as illustrated in Figure~\ref{fig:mlp_cnn_chnnet}\subref{fig:img4}.

In both experiments, we used a learning rate of $1\times10^{-3}$, batch size of 128, and sparse categorical crossentropy loss with logits disabled. ReLU nonlinearities activated the hidden layers, with softmax applied at the output to produce class probabilities for the 10-class classification task. This consistent training regimen, implemented in TensorFlow, facilitated a controlled and rigorous comparison of the models on MNIST.

\subsection{Experiment with CNN Architectures}
\subsubsection{Convolutional Adaptation of CHNNet}
Consider that $x \in \mathbb{R}^{h \times w \times c_{in}}$ denotes the input feature map, where $h$ and $w$ represent the height and width of the input feature map and $c_{in}$ is the number of input channels. For the $l^{th}$ layer, the model first computes an intermediate feature map as follows:
\begin{equation}
    \label{cnn1}
    h_i^{[l]} = x^{[l]} \ast K_i^{[l]}; ~K_i^{[l]} \in \mathbb{R}^{k_h^{[l]} \times k_w^{[l]} \times c_{in}^{[l]} \times c_{out}^{[l]}}
\end{equation}
Here, $k_h$ and $k_w$ represent the height and width of the kernel, while $c_{in}$ and $c_{out}$ denote the number of input and output channels, respectively. In fact, equation (\ref{cnn1}) corresponds to the identity mapping as described in equation (\ref{inf2}). Subsequently, The model computes the following:
\begin{equation}
    \label{cnn2}
    h_h^{[l]} = h_i^{[l]} \ast K_h^{[l]} ; ~K_h^{[l]} \in \mathbb{R}^{\tilde{k}_h^{[l]} \times \tilde{k}_w^{[l]} \times c_{out}^{[l]} \times c_{out}^{[l]}}
\end{equation}

When the spatial kernels are shared across channels, $\tilde{k}_h$ and $\tilde{k}_w$ are equal to $k_h$ and $k_w$, respectively. Conversely, when kernel sharing is not applied, $\tilde{k}_h$ and $\tilde{k}_w$ are equal to 1. 

Subsequently, the pre-activation output is computed as follows:
\begin{equation}
    \label{cnn3}
    z^{[l]} = h_i^{[l]} + h_{h}^{[l]}
\end{equation}

This formulation aligns with the pre-activation computation for hidden layers in equation (\ref{inf1}) under the identification \(a^{[l-1]}\!\equiv x^{[l]}\), \(h^{[l]}\!\equiv h_i^{[l]}\), \(W_{1}^{[l]}\!\leftrightarrow K_{i}^{[l]}\), \(W_{2}^{[l]}\!\leftrightarrow K_{h}^{[l]}\), and \(b^{[l]}=\vec{0}\).

In the linear case with unit hidden stride, the composition reduces to a single effective kernel $K_e = K_i + (K_i \ast K_h)$. This is consistent with the one-layer reduction presented in equation (\ref{lit3}) and emphasizes that the intra-layer term of CHNNet represents a same-step coupling.

\subsubsection{Experiments}
We evaluated the convergence behavior of CHNNet against a conventional CNN architecture on CIFAR-10. In the unequal-parameter setting, both models consisted of three stacked convolutional stages with 5 $\times$ 5 kernels and same padding, featuring channel widths of 64, 128 and 256. Each stage is followed by batch normalization, a rectified linear unit activation, and a 3 $\times$ 3 max pooling with stride 2 and valid padding. The classifier head flattened the feature representation and applied a fully connected layer with 1000 units, followed by batch normalization, rectified linear activation, and 10-way softmax output. The baseline CNN comprised 3.3 million parameters, whereas the CHNNet variant, which introduces intra-layer channel mixing at each stage while maintaining the same depth and pooling structure, contained 4.4 million parameters. The loss curve of CHNNet in Figure~\ref{fig:mlp_cnn_chnnet}\subref{fig:img5} show an overall downward trend, with a transient spike around epoch 55 before resuming their decline. This behavior is presumably due to moment-based update volatility on an atypical mini-batch interacting with batch normalization updates late in training, temporarily perturbing the running statistics before they realign.

To provide a balanced comparison, we also conducted experiments with a matched number of parameters. The conventional CNN was widened to 90, 150 and 280 channels followed by a fully connected head with 1250 units, resulting in 4.5 million parameters, whereas CHNNet retained 64, 128 and 256 channels with a 1000-unit CHN head. Despite this near parity, CHNNet retains its convergence advantage, as visualized in Figure~\ref{fig:mlp_cnn_chnnet}\subref{fig:img6}.

In the experiments, the inputs were standardized per channel and the models were trained using mini-batches of size 128 and the Adam optimizer with $\beta_1=0.9$, $\beta_2=0.999$, and $\varepsilon=10^{-8}$ at a learning rate of $1\times10^{-4}$. Both the unequal-parameter and matched-parameter experiments were run for 70 epochs under the same training schedule and evaluation settings.

\subsection{Discussion}
CHNNet consistently outperformed the baseline architectures in both fully connected and convolutional settings, as illustrated in Figure~\ref{fig:mlp_cnn_chnnet}. As reported in Table~\ref{tab:mlp_chn_params_equal_unequal}, CHNNet achieved slightly higher test accuracy and lower loss than the MLP, demonstrating the effectiveness of its intra-layer connection mechanism. Similarly, Table~\ref{tab:cnn_chn_params_equal_unequal} shows that CHNNet surpassed the CNN in accuracy while exhibiting reduced loss, indicating that intra-layer mixing strengthens convolutional feature extraction.

Evaluation under equal-parameter configurations provides additional insight into the source of these improvements. In both MLP and CNN settings with increased parameters, CHNNet converged more rapidly and consistently attained higher accuracy with lower loss than the baseline models. These results suggest that the performance gains stem primarily from the architectural advantages introduced by intra-layer connections, rather than from the differences in parameter count.

\begin{table}[t]
\centering
\caption{Comparison of MLP and CHNNet.}
\label{tab:mlp_chn_params_equal_unequal}
\begin{tabular}{l c c c}
\toprule
Model & Total Params & Accuracy & Loss \\
\midrule
MLP     & 1.7M & 0.9836 & 0.1509 \\
CHNNet  & 3.7M & 0.9860 & 0.1208 \\
\midrule
MLP     & 3.7M & 0.9847 & 0.1309 \\
CHNNet  & 3.7M & 0.9870 & 0.1163 \\
\bottomrule
\end{tabular}
\end{table}

\begin{table}[t]
\centering
\caption{Comparison of CNN and CHNNet.}
\label{tab:cnn_chn_params_equal_unequal}
\begin{tabular}{l c c c}
\toprule
Model & Total Params & Accuracy & Loss \\
\midrule
CNN     & 3.3M & 0.7122 & 2.2610 \\
CHNNet  & 4.4M & 0.7941 & 1.0325 \\
\midrule
CNN     & 4.5M & 0.7452 & 1.9401 \\
CHNNet  & 4.4M & 0.7747 & 1.1329 \\
\bottomrule
\end{tabular}
\end{table}

\section{Future Work}
While CHNNet demonstrates the potential benefits of incorporating intra-layer connectivity in feedforward networks, several promising directions merit further study:
\begin{itemize}
    \item While CHNNet presents one approach to intra-layer connectivity, exploring alternative patterns or weighting strategies for hidden neuron interactions may uncover more efficient or stable configurations that further enhance the performance of the network.
    \item CHNNet has thus far been evaluated within the MLP and CNN architectures. However, its applicability could be further examined across other paradigms such as Recurrent Neural Networks (RNNs), Spiking Neural Networks (SNNs), Variational Autoencoders (VAEs), and Transformers, where introducing intra-layer connectivity may have significant implications for learning dynamics.
    \item CHNNet has so far been evaluated exclusively using the backpropagation algorithm. Its compatibility and performance with alternative learning methods, such as the forward-forward algorithm~\cite{hin22} or predictive coding~\cite{rao99}, remain to be explored.
    \item The effects of hyperparameters such as activation functions, optimization algorithms, and other training configurations on CHNNet’s performance are yet to be rigorously studied.
\end{itemize}

\section{Conclusion}
In this study, we investigate the impact of introducing intra-layer connectivity among hidden neurons within conventional feedforward architectures. The principal contributions of this work are summarized as follows:
\begin{itemize}
  \item We introduce CHNNet, a biologically plausible network with intra-layer connections among hidden neurons, facilitating efficient information sharing within each hidden layer.
  \item We formulate equations for hidden layer activations for the inference process and adapt the backpropagation algorithm to compute gradients accordingly during the learning process.
  \item Through theoretical analysis of the model, we demonstrate that CHNNet achieves faster convergence compared to conventional FNNs.
  \item We have evaluated the proposed model on benchmark datasets and empirically demonstrated a substantial improvement in convergence rate compared to conventional FNNs.
\end{itemize}

The study underscores that even minimal structural modifications, specifically introducing interactions among hidden neurons, can enhance the learning behavior of standard architectures. This work points toward a broader reconsideration of connectivity patterns in neural networks, suggesting that gains in efficiency and stability may arise not only from increased depth or complexity, but also from fostering richer intra-layer coordination.

{
    \small
    \bibliographystyle{ieeenat_fullname}
    \bibliography{main}
}


\end{document}